\documentclass[runningheads]{llncs}

\usepackage[T1]{fontenc}
\usepackage{graphicx}
\usepackage{amsmath,amssymb,amsfonts}
\usepackage{booktabs}
\usepackage{multirow}
\usepackage{tabularx}
\usepackage{algorithm}
\usepackage{algpseudocode}
\usepackage{xcolor}
\usepackage{float}
\usepackage{pgfplots}
\usepackage{svg}
\usepackage{wrapfig}
\usepackage{enumitem}
\setlist{nosep}
\usepackage{subcaption}
\usepackage[hidelinks]{hyperref}

\usepackage{orcidlink}

\newcommand{\revise}[1]{#1}

\newcommand{\camera}[1]{#1}

\newcommand{\parag}[1]{\vspace{-3mm}\paragraph{#1}}

\newcommand{\method}{CoreCalib}

\makeatletter
\newcommand{\printfnsymbol}[1]{%
  \textsuperscript{\@fnsymbol{#1}}%
}
\makeatother

\begin{document}
\title{Density-Aware Calibration for Resource-Efficient Medical Vision-Language Models}

\titlerunning{Density-Aware Calibration for Resource-Efficient Medical VLMs}

\author{
Tien-Dat Dam\inst{1,2}\orcidlink{0009-0009-9507-2392}\and
Huu-Thuan Huynh\inst{1,2}\orcidlink{0009-0003-3984-3355}
\and
Minh-Triet Tran\inst{1,2}\orcidlink{0000-0003-3046-3041}
\and
Trung-Nghia Le\thanks{Corresponding author.}\inst{1,2}\orcidlink{0000-0002-7363-2610}
}

\authorrunning{Tien-Dat Dam et al.}

\institute{
University of Science, Ho Chi Minh City, Vietnam
\and
Vietnam National University, Ho Chi Minh City, Vietnam
\\
\email{
23120118@student.hcmus.edu.vn,  hhthuan@hcmus.edu.vn, 
tmtriet@fit.hcmus.edu.vn, 
ltnghia@fit.hcmus.edu.vn
}
}

\maketitle              
\begin{abstract}
\revise{
Medical vision-language models (VLMs) show promise for clinical image understanding, but their computational and memory requirements hinder deployment on resource-constrained clinical hardware. Post-training quantization (PTQ) offers a practical compression strategy without retraining, yet its effectiveness depends strongly on the calibration data. We propose CoreCalib, a training-free and quantizer-agnostic framework for constructing compact calibration sets that preserve the target medical VQA distribution. CoreCalib stratifies data by clinical question class, identifies representative visual modes using frozen DINOv2 features with UMAP and HDBSCAN, and expands selected images into class-balanced question-answer pairs. On Kvasir-VQA-x1, we evaluate CoreCalib with MASQuant, GPTQ, and QIG using a fixed 128-pair budget. CoreCalib achieves the best METEOR with all three quantizers and the best ROUGE-L with MASQuant and GPTQ. With W4A8 MASQuant, it improves METEOR from 0.5672 with COCO calibration to 0.6234, recovering approximately 41\% of the degradation relative to BF16. These results support CoreCalib as a data-centric component for resource-efficient medical VLM deployment. Project page: \url{https://magnusdtd.github.io/CoreCalib/}.
}

\end{abstract}
\section{Introduction}


\revise{Medical visual question answering (VQA) aims to answer clinical questions about medical images and has become a promising building block for decision support in radiology, pathology, and gastrointestinal (GI) endoscopy~\cite{lau2018dataset,he2020pathvqa,gautam2024kvasir}. Modern VQA systems increasingly rely on large vision-language models (VLMs), such as Qwen3.5~\cite{qwen3.5}, that can be adapted to clinical domains with parameter-efficient fine-tuning~\cite{hu2021lora,dettmers2023qlora}. On the Kvasir-VQA-x1 benchmark~\cite{gautam2025kvasir}, a 4-billion-parameter Qwen3.5 model fine-tuned for GI endoscopy VQA provides strong performance. However, deploying such a model in a clinical environment is substantially different from running it in a research setting: inference close to the endoscopy suite favors local and on-premises computing, where accelerator memory, latency, and computational budgets can be considerably tighter than those of large-scale research servers. A medical VLM that requires substantial resources can therefore become difficult to deploy beyond high-end GPU infrastructure.}


\revise{A practical clinical VLM should ideally operate across a spectrum of computing environments, ranging from local GPU servers and hospital workstations to resource-constrained computers and, ultimately, edge systems located close to image acquisition. Model compression is an important step toward this goal. Post-training quantization (PTQ) is particularly attractive because it converts a trained model to low-bit representations without requiring expensive retraining~\cite{nagel2021white}. By reducing the numerical precision of weights and, for suitable methods, activations, PTQ can reduce the resource burden of inference while preserving the original model architecture and learned capabilities. This makes PTQ a natural interface between a trained medical VLM and a resource-efficient clinical deployment pipeline.}

\revise{However, low-bit quantization introduces a fundamental trade-off between resource efficiency and model fidelity. This trade-off is especially consequential for medical VLMs, where changes in visual-language representations can alter clinically meaningful answers. Recent PTQ methods for multimodal models differ in how they estimate and control quantization error: SmoothQuant-style methods migrate activation outliers into weights~\cite{xiao2023smoothquant,li2025mbq,hu2026masquant}, GPTQ reconstructs weights using approximate second-order information~\cite{frantar2022gptq}, and QIG reweights tokens using quantization-aware integrated gradients~\cite{xiang2026fine}. Despite these advances, the calibration data used by PTQ is often treated as a fixed implementation detail. Multimodal PTQ methods commonly use small generic COCO-based image-caption sets~\cite{li2025mbq,xiang2026fine}, even when the target model has been specialized to a substantially different clinical distribution.}

\revise{For medical VQA, this calibration mismatch is particularly important because a calibration example is not simply an image. The same endoscopic image may be paired with questions about abnormality presence, polyp type, instrument presence, counts, anatomical landmarks, or image quality. These question types induce different textual and multimodal interactions, while the underlying images exhibit recurrent but fine-grained visual patterns. Consequently, a calibration set can be visually representative but poorly cover the clinical tasks that the model must perform, or it can cover frequent question classes while spending a limited visual budget on atypical images. We therefore view calibration as a multimodal workload-construction problem: the calibration set should jointly represent the target visual distribution and the clinical question distribution under a very small budget.}


\revise{Our observations on Kvasir-VQA-x1 further motivate this view. Simply replacing generic calibration data with randomly selected in-domain examples is insufficient: under W4A8 MASQuant~\cite{hu2026masquant}, a random in-domain calibration set produces a 0.0870 METEOR drop relative to the half-precision model. This indicates that the path from a full-precision medical VLM toward a resource-efficient model is sensitive not only to the final quantization precision, but also to which target-domain examples are exposed to the PTQ procedure.}


\revise{To address this problem, we propose \textbf{Core-Distribution Calibration (\method{})}, a training-free and quantizer-agnostic framework designed as a data-centric component of medical VLM compression. \method{} explicitly separates three decisions that are often conflated: which clinical question classes should contribute to calibration, which real images best represent the visual modes of each class, and which QA pairs should be drawn from those images. The first stage stratifies the training data by clinical question class. The second embeds images with a frozen DINOv2 encoder~\cite{oquab2023dinov2}, reduces the representation with UMAP~\cite{mcinnes2018umap}, and uses HDBSCAN~\cite{campello2013density} with CoDA-style post-processing~\cite{zhou2026coda} to retrieve real representatives from high-density regions. The third expands the representatives into their available QA pairs and fills the fixed calibration budget with a class-balanced round-robin policy. The resulting calibration set is passed unchanged to an existing PTQ algorithm. Unlike approaches that introduce a new learned compression component, \method{} does not modify the fine-tuned VLM or the internal objective of the PTQ algorithm. Instead, it improves the information presented to the existing quantization procedure. This separation makes the method compatible with different PTQ mechanisms and allows calibration to be optimized independently of the final deployment hardware.}


\revise{We evaluate \method{} on Kvasir-VQA-x1 with three PTQ methods, MASQuant, GPTQ, and QIG, using the same 128-pair calibration budget. \method{} obtains the best METEOR among the compared calibration strategies for all three quantizers, and the best ROUGE-L with MASQuant and GPTQ. Under W4A8 MASQuant, it improves METEOR from 0.5672 with generic COCO calibration to 0.6234, recovering approximately 41\% of the METEOR degradation relative to the BF16 model. These results show that calibration-set construction is a substantive component of capability-preserving medical VLM compression and provides a practical data-centric step toward resource-constrained clinical deployment.}

\revise{Our contributions are as follows:}

\begin{itemize}
    \item \revise{We identify calibration-data construction as an important component of resource-efficient medical VLM deployment and formulate it as a fixed-budget data selection problem for post-training quantization.}

    \item \revise{We propose CoreCalib, a training-free and quantizer-agnostic framework that combines clinical question-class stratification, density-aware visual core discovery, and class-balanced question-answer expansion to construct a compact calibration set from the target medical distribution.}

    \item \revise{We systematically evaluate CoreCalib with three representative PTQ methods, namely MASQuant, GPTQ, and QIG, under the same 128-pair calibration budget. CoreCalib achieves the best METEOR with all three quantizers and the best ROUGE-L with two of the three, demonstrating robust capability retention across different quantization mechanisms.}

\end{itemize}

\section{Related Work}

\parag{\textbf{Post-training quantization of LLMs and VLMs.}}

\revise{PTQ quantizes a pretrained model using a small calibration set instead of retraining~\cite{nagel2021white}. For large language models, GPTQ~\cite{frantar2022gptq} quantizes weights column by column and compensates for reconstruction error using an approximate Hessian, while AWQ~\cite{lin2024awq} protects salient weight channels identified from activation magnitudes. For activation quantization, SmoothQuant~\cite{xiao2023smoothquant} migrates outliers from activations to weights, OmniQuant~\cite{shao2024omniquant} learns quantization factors, and QuaRot~\cite{ashkboos2024quarot} removes outliers with rotations. Multimodal models introduce an additional modality gap: MBQ~\cite{li2025mbq} reweights reconstruction according to modality sensitivity, MASQuant~\cite{hu2026masquant} learns modality-specific smoothing factors and compensates for modality differences with a low-rank term, and QIG~\cite{xiang2026fine} refines modality-level sensitivity to token-level importance with integrated gradients. These methods primarily improve the quantizer itself. \method{} is complementary: the quantizer remains unchanged, while the calibration workload is redesigned for the target clinical distribution.}

\parag{\textbf{Calibration data for compression.}} 

\revise{The source and composition of calibration data can affect downstream behavior after compression. Williams and Aletras~\cite{williams2024impact} showed that calibration-data choice changes the performance of quantized and pruned LLMs, while Ji et al.~\cite{ji2025beware} investigated the relation between calibration distribution and pruning quality. COLA~\cite{he2026preserving} curates calibration data per capability and selects samples close to K-Means centroids of projected activations, while DPQ~\cite{yang2026target} selects low-margin examples to preserve uncertainty. Data-free methods instead synthesize calibration inputs from internal statistics or generative models~\cite{cai2020zeroq,li2024genq}. These studies motivate matching calibration inputs to the target workload, but medical VQA has a distinctive structure: one image can support multiple QA pairs spanning heterogeneous clinical tasks. \method{} explicitly exploits this structure by selecting representative images first and then expanding them into balanced multimodal interactions.}

\parag{\textbf{Dataset distillation and density-based core selection.}}

\revise{Dataset distillation compresses a training set into a few synthetic samples that preserve useful information~\cite{wang2018dataset}, while core-set selection chooses real samples that efficiently cover a distribution~\cite{sener2017active}. Diffusion-based distillation methods such as D$^4$M~\cite{su2024d} and MGD$^3$~\cite{chan2025mgd} use K-Means prototypes in latent spaces to guide generation. CoDA~\cite{zhou2026coda} instead identifies an intrinsic core distribution using UMAP~\cite{mcinnes2018umap} and HDBSCAN~\cite{campello2013density}, followed by post-processing to obtain a prescribed number of representatives. \method{} transfers this density-based discovery principle from dataset distillation to PTQ calibration, but retains real target-domain images and additionally introduces clinical question-class stratification and QA-level balancing. The resulting objective is not dataset distillation itself, but construction of a compact multimodal workload for capability-preserving quantization.}

\section{Proposed Method}

\subsection{Overview}

\begin{figure}[t!]
    \centering
    \includegraphics[width=\linewidth]{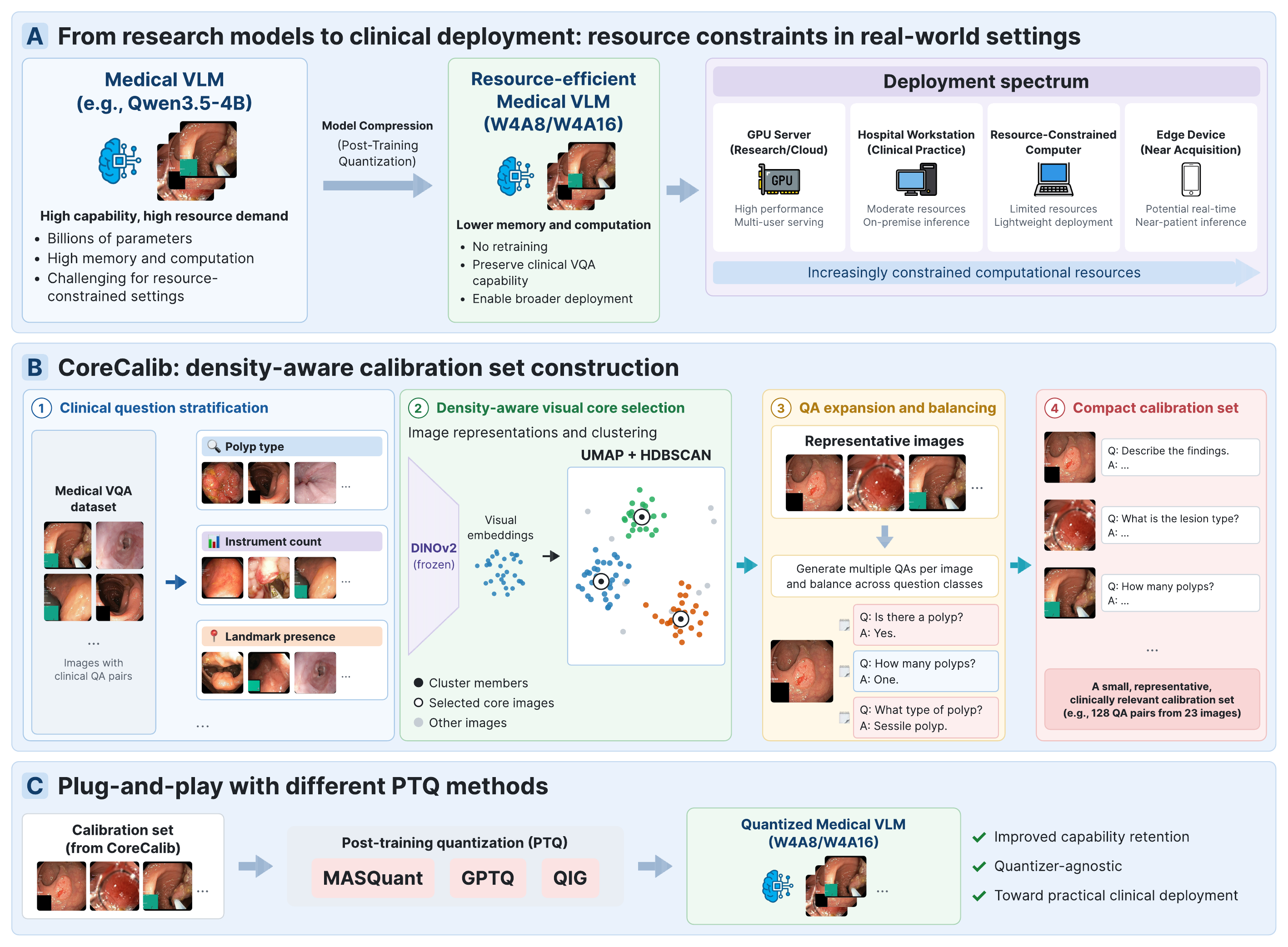}
    \caption{\revise{Overview of CoreCalib and its role in resource-efficient medical VLM deployment. (A) A medical VLM provides strong clinical VQA capability but requires substantial computational resources; PTQ converts it into a low-bit model intended to reduce the resource burden across a spectrum of computing environments. (B) CoreCalib constructs a compact calibration workload by (1) stratifying the medical VQA data by clinical question class, (2) selecting density-aware representative images using frozen DINOv2 features with UMAP and HDBSCAN, (3) expanding representative images into multiple QA pairs and balancing them across question classes, and (4) forming a small calibration set, such as 128 QA pairs from 23 images. (C) The calibration set can be used with different PTQ methods, including MASQuant, GPTQ, and QIG, to produce a quantized medical VLM with improved capability retention.}}
    \label{fig:main_pipeline}
\end{figure}

\revise{\method{} is designed as a plug-in calibration stage between a fine-tuned medical VLM and an existing PTQ algorithm (Fig.~\ref{fig:main_pipeline}). The framework has three data-centric stages: (1) clinical question-class stratification, which ensures that the calibration workload covers heterogeneous medical VQA tasks; (2) density-aware visual core selection, which identifies representative real images in high-density regions of each class-specific visual distribution; and (3) QA expansion and class-balanced sampling, which converts the selected images into a fixed-size multimodal calibration workload. The final set is then consumed by an unmodified PTQ method. This design separates calibration-set construction from quantizer design and therefore supports different PTQ mechanisms and deployment targets.}

\subsection{\revise{Calibration Problem Formulation}}

\revise{Let $\mathcal{D}=\{(I_i,q_i,a_i,c_i)\}$ denote the medical VQA training set, where $I_i$ is an image, $q_i$ is a question, $a_i$ is an answer, and $c_i\in\mathcal{C}$ is its clinical question class. A PTQ algorithm $\mathcal{Q}$ receives a calibration set $\mathcal{S}$ of fixed size $N$ and produces a quantized model from a fine-tuned model $f$. Conventional pipelines treat $\mathcal{S}$ as an externally provided set; \method{} instead constructs:
\begin{equation}
\mathcal{S}=\mathcal{F}(\mathcal{D},N), \qquad |\mathcal{S}|=N,
\end{equation}
where $\mathcal{F}$ is independent of the internal optimization of $\mathcal{Q}$. The practical objective is to maximize useful coverage under the strict calibration budget: every clinical question class should contribute to the calibration workload, while the selected images within each class should represent dense modes of the target visual distribution. This formulation makes calibration-set construction an explicit component of resource-efficient medical VLM compression rather than a fixed preprocessing choice.}

\subsection{Question-Class Stratification}

Unlike image classification datasets, a VQA training set has no single label per image: the same endoscopic image is paired with questions about polyps, instruments, anatomical landmarks, and image quality. The answers to these questions follow very different formats, from yes/no statements to counts and clinical entity names, and each format exercises different parts of the language model. \camera{We use the question classes $C$ provided by each dataset as strata}~\cite{gautam2025kvasir,lau2018dataset}. For each class $c \in C$, we define the subset $D_c = \{(I, q, a, c') \in D \mid c' = c\}$ and its image set $I_c$. Since an image can be associated with several classes, the strata overlap at the image level, allowing a single representative image to serve multiple question types.

\subsection{Core Distribution Discovery}

\revise{Within each stratum, we retrieve $K$ real images that summarize the dominant visual modes of $I_c$. Following CoDA~\cite{zhou2026coda}, we use density-based discovery rather than centroid-based clustering because the calibration budget is small and medical image distributions may contain heterogeneous, non-convex, or unevenly populated modes. Each image is encoded by a frozen, self-supervised DINOv2 encoder $\phi$, which was not trained on the target dataset. \camera{Let $\psi$ denote the UMAP dimensionality-reduction mapping. It} maps the features to $\mathbf{z}_I=\psi(\phi(I))$, preserving local neighborhoods in a compact space. HDBSCAN is then applied to $\{\mathbf{z}_I\}_{I\in I_c}$, producing $M_c$ clusters $\{G_k\}$ and a membership probability $p(I)$ for each clustered image, while low-density points can be labeled as noise. For each cluster, we select the real image with the highest membership probability:
\begin{equation}
  r_k=\arg\max_{I\in G_k}p(I).
\end{equation}
}
\revise{Because $M_c$ is data-dependent, we adopt CoDA's post-processing to obtain exactly $K$ representatives: if $M_c\ge K$, representatives from the $K$ largest clusters are retained; otherwise, large clusters are recursively split, outliers are clustered with K-Means, and, if necessary, the minimum cluster size is relaxed. This yields an ordered list $R_c=[r_1,\dots,r_K]$. The resulting representatives are actual medical images located in dense regions of the target distribution, reducing the likelihood that a limited calibration budget is spent on atypical cases.}

\subsection{Question-Answer Expansion and Class-Balanced Sampling}

\revise{PTQ algorithms consume complete multimodal inputs, so representative images must be paired with text. For each class $c$, we collect the list $L_c$ of all QA pairs in $D_c$ whose image belongs to $R_c$, ordered according to the rank of the representative image. The calibration set is filled in round-robin order over classes: at each round, the next available QA pair from every non-empty $L_c$ is appended until $|S|=N$. This policy prevents frequent question classes from consuming the entire budget while retaining multiple clinical interactions associated with a representative image. \camera{In the Kvasir-VQA-x1 setting}, 128 QA pairs are obtained from only 23 unique images. Thus, the calibration budget represents both multimodal interaction coverage and a compact visual footprint, which is useful when the quantization procedure must be performed with limited access to clinical data.}

\section{Experiments}

\subsection{Implementation Details}

\revise{The half-precision model is Qwen3.5-4B~\cite{qwen3.5} fine-tuned on the Kvasir-VQA-x1 training split with QLoRA~\cite{dettmers2023qlora}, using LoRA~\cite{hu2021lora} with rank 16 and $\alpha=32$ on the attention, MLP, and vision-language projection layers. We apply three PTQ methods using their official implementations. MASQuant~\cite{hu2026masquant} quantizes weights to 4 bits and activations to 8 bits (W4A8) with modality-specific smoothing factors optimized for 2 epochs with batch size 1 and a low-rank compensation ratio of 0.05. GPTQ~\cite{frantar2022gptq} quantizes weights to 4 bits with group size 128 and dampening ratio 0.01. QIG~\cite{xiang2026fine} quantizes the model to W4A8 with integrated-gradient token reweighting, distortion-aware scale search, and a mean absolute error loss. Every calibration strategy uses exactly $N=128$ samples, so differences in downstream performance are attributable to calibration construction rather than calibration-set size.}

\revise{For \method{}, image features are extracted with DINOv2 ViT-B/14~\cite{oquab2023dinov2}. UMAP uses 50 neighbors, 50 output dimensions, and a minimum distance of 0. HDBSCAN uses a minimum cluster size of 25 and 3 minimum samples, and $K=10$ representatives are retrieved per class. Metrics are computed with the Hugging Face implementations. All experiments run on NVIDIA A100 40GB GPUs. The calibration procedure itself does not modify the fine-tuned model or the PTQ objectives.}

\subsection{Dataset and Metrics}

\camera{Our primary experiments use Kvasir-VQA-x1}~\cite{gautam2025kvasir}, a GI endoscopy VQA dataset of 159,549 QA pairs built on 6,500 images from HyperKvasir~\cite{borgli2020hyperkvasir} and Kvasir-Instrument~\cite{jha2021kvasir}. Questions are annotated with 18 clinical question classes (\textit{e.g.}, polyp type, instrument count, landmark presence). \camera{For this dataset, in-domain calibration sets were drawn only from the training split, and VQA quality was evaluated on the full test split of 15,955 samples.} Following the benchmark protocol, we report BLEU~\cite{papineni-etal-2002-bleu}, ROUGE-L~\cite{lin-2004-rouge}, and METEOR~\cite{banerjee-lavie-2005-meteor}. \camera{Additional evaluations cover VQA-RAD~\cite{lau2018dataset} dataset.}

\subsection{Comparisons with Baselines}

\revise{We compare \method{} with three calibration strategies under the same 128-pair budget and the same QA format. COCO uses 128 ShareGPT4V image-caption pairs~\cite{chen2024sharegpt4v} and represents the generic calibration setting used by multimodal PTQ methods~\cite{li2025mbq,xiang2026fine}. Random samples training images uniformly at random and applies the same QA expansion and class-balanced sampling procedure as \method{}, isolating the effect of representative image selection. K-Means replaces density-based discovery with per-class K-Means ($k=K$) on the same DINOv2 features and selects the real image nearest to each centroid, following prototype selection in prior calibration and distillation work~\cite{he2026preserving,chan2025mgd}.}

\begin{table}[t!]
  \centering
  \caption{Comparison of calibration strategies on the Kvasir-VQA-x1 test set. Within each quantizer, best and second-best results are shown in \textbf{bold} and \underline{underline}.}
  \label{tab:main}
  {\small
    \setlength{\tabcolsep}{4pt}
    \begin{tabular}{llccc}
      \toprule
      \textbf{Quantizer} & \textbf{Calibration} & \textbf{BLEU} $\uparrow$ & \textbf{ROUGE-L} $\uparrow$ & \textbf{METEOR} $\uparrow$ \\
      \midrule
      \multicolumn{2}{l}{\textit{Half precision (BF16)}} & 0.4964 & 0.7005 & 0.7040 \\
      \midrule
      \multirow{4}{*}{\shortstack[l]{MASQuant \\ (W4A8)}}
      & COCO             & 0.3313            & 0.5486            & 0.5672            \\
      & Random           & 0.3827 & 0.5991 & \underline{0.6170} \\
      & K-Means          & \textbf{0.3954}   & \underline{0.6063} & \underline{0.6170}           \\
      & \method{} (Ours) & \underline{0.3943} & \textbf{0.6103}   & \textbf{0.6234}   \\
      \midrule
      \multirow{4}{*}{\shortstack[l]{GPTQ \\ (W4A16)}}
      & COCO             & 0.4172            & 0.6373            & 0.6545            \\
      & Random           & \underline{0.4438}   & \underline{0.6586}   & 0.6625   \\
      & K-Means          & 0.4380            & 0.6554            & \underline{0.6642} \\
      & \method{} (Ours) & \textbf{0.4473} & \textbf{0.6608} & \textbf{0.6649}           \\
      \midrule
      \multirow{4}{*}{\shortstack[l]{QIG \\ (W4A8)}}
      & COCO             & \textbf{0.3099}   & 0.5310 & 0.5513 \\
      & Random           & 0.3041           & \textbf{0.5321}           & \underline{0.5565}          \\
      & K-Means          & 0.3048            & 0.5235            & 0.5494            \\
      & \method{} (Ours) & \underline{0.3054} & \underline{0.5313}   & \textbf{0.5581}   \\
      \bottomrule
    \end{tabular}
  }
  \vspace{-6mm}
\end{table}

\paragraph{\textbf{Quantitative Results.}}

Tab.~\ref{tab:main} shows that the calibration data had a large effect on quantized quality. Generic COCO calibration caused the largest degradation, and every in-domain strategy recovered a substantial part of it, most notably with MASQuant, where \method{} improved METEOR by 0.0562 over COCO. Among in-domain strategies, \method{} achieved the best METEOR with every quantizer and the best ROUGE-L with MASQuant and GPTQ, while BLEU was occasionally slightly higher with K-Means or COCO. GPTQ and QIG were less sensitive to the calibration data: with QIG, \method{}
ranked first on METEOR and second on ROUGE-L, 0.0008 behind Random, which indicates that its benefit is not tied to a specific PTQ algorithm. \camera{Due to computational resource constraints, results are reported as single-run point estimates without multi-seed evaluation.}

\begin{figure}[t!]
  \centering
  \includegraphics[width=0.8\linewidth]{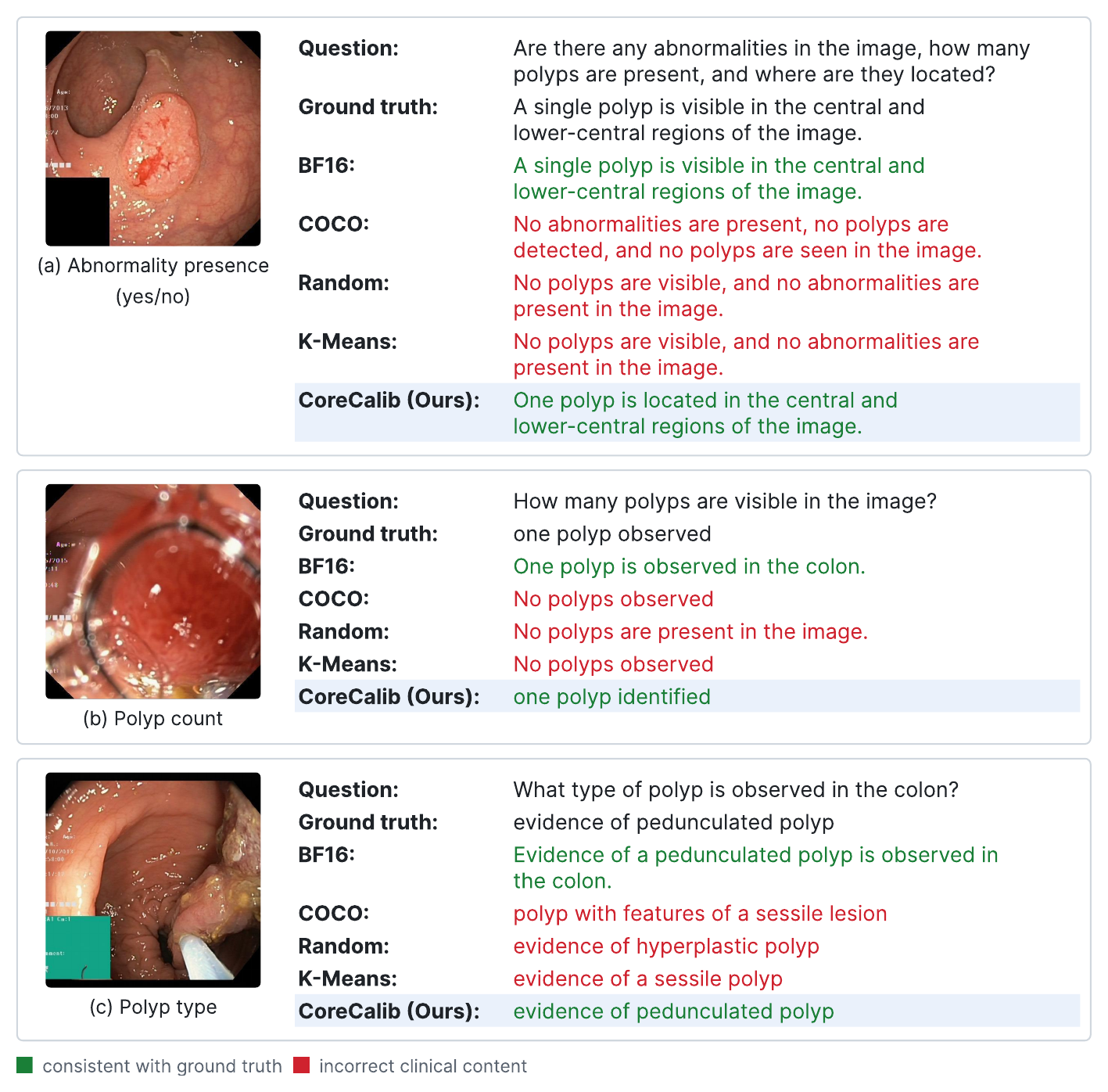}
  \caption{Qualitative examples on the Kvasir-VQA-x1 test set with MASQuant (W4A8) under different calibration sets. Green and red mark answers that are consistent with and contradict the ground truth, respectively.}
  \label{fig:qualitative}
\end{figure}

\paragraph{\textbf{Qualitative Results.}}
\revise{Fig.~\ref{fig:qualitative} shows representative test examples with MASQuant (W4A8). In the abnormality-presence and polyp-count examples, COCO, Random, and K-Means calibration lead the quantized model to deny a visible polyp, whereas \method{} preserves the finding and its location consistently with the half-precision model and ground truth. In the polyp-type example, the baselines describe a pedunculated polyp as sessile or hyperplastic, whereas \method{} reproduces the reference type. These examples illustrate clinically meaningful content errors-missed abnormality, incorrect lesion count, and incorrect lesion type-that can arise after compression. They should not be interpreted as clinical validation, but they provide qualitative evidence that representative target-domain calibration can preserve fine-grained clinical findings.}

\subsection{Ablation Study}

\revise{Tab.~\ref{tab:ablation} isolates the components of \method{} with MASQuant. Replacing density-based core discovery with Random or K-Means selection reduces ROUGE-L and METEOR, indicating that the gain is not explained solely by the DINOv2 feature space. Removing question-class stratification also reduces the overall performance, supporting the need to cover heterogeneous clinical tasks explicitly. Selecting the second-most-confident member of each cluster increases BLEU but decreases ROUGE-L and METEOR, suggesting that points located closest to the density core are preferable for the main evaluation metrics. The complete \method{} configuration achieves the best ROUGE-L and METEOR, consistent with the design principle that visual representativeness and clinical-task coverage should be considered jointly.}

\begin{table}[t!]
  \centering
  \caption{Ablation of \method{} components with MASQuant on Kvasir-VQA-x1. Strat.: question-class stratification; Discovery: how representative images are selected; Rank: which representative of each cluster is used.}
  \label{tab:ablation}
  {\small
    \setlength{\tabcolsep}{2pt}
    \begin{tabular}{lcccccc}
      \toprule
      \textbf{Variant}   & \textbf{Strat.} & \textbf{Discovery} & \textbf{Rank} & \textbf{BLEU} $\uparrow$ & \textbf{ROUGE-L} $\uparrow$ & \textbf{METEOR} $\uparrow$ \\
      \midrule
      Random             & \checkmark      & --                 & --            & 0.3827                   & 0.5991                      & 0.6170                     \\
      K-Means            & \checkmark      & K-Means            & 1st           & \underline{0.3954}       & \underline{0.6063}          & 0.6170                     \\
      w/o stratification & $\times$        & HDBSCAN            & 1st           & 0.3905                   & 0.6029                      & \underline{0.6190}         \\
      2nd member         & \checkmark      & HDBSCAN            & 2nd           & \textbf{0.4002}          & 0.6014                      & 0.6173                     \\
      \method{}          & \checkmark      & HDBSCAN            & 1st           & 0.3943                   & \textbf{0.6103}             & \textbf{0.6234}            \\
      \bottomrule
    \end{tabular}
  }
  \vspace{-6mm}
\end{table}

\revise{Tab.~\ref{tab:budget} reports the effect of the calibration budget $N$. BLEU and ROUGE-L increase from 32 to 256 samples, whereas METEOR is non-monotonic and is already strong at $N=64$. At $N=128$, \method{} uses only 23 unique images while providing a strong balance across the three metrics. Increasing the budget to 256 samples improves BLEU and ROUGE-L but requires 35 unique images, while METEOR decreases relative to $N=64$. This behavior indicates that simply increasing calibration-set size does not guarantee uniformly better capability retention and motivates selecting representative samples under a fixed resource and data budget.}

\begin{table}[t!]
  \centering
  \caption{Effect of the calibration budget $N$ on \method{} with MASQuant (W4A8).}
  \label{tab:budget}
  {\small
    \begin{tabular}{ccccc}
      \toprule
      \textbf{N} & \textbf{\#Images} & \textbf{BLEU} $\uparrow$ & \textbf{ROUGE-L} $\uparrow$ & \textbf{METEOR} $\uparrow$ \\
      \midrule
      32  & 12                & 0.3851                   & 0.6054                      & 0.6184                     \\
      64  & 16                & 0.3891                   & 0.6066                      & \textbf{0.6326}            \\
      128 & 23                & \underline{0.3943}       & \underline{0.6103}          & 0.6234                     \\
      256 & 35                & \textbf{0.3962}          & \textbf{0.6155}             & \underline{0.6277}         \\
      \bottomrule
    \end{tabular}
  }
  \vspace{-6mm}
\end{table}

\camera{To examine cross-modality generalization beyond GI endoscopy, Tab.~\ref{tab:vqarad} evaluates GPTQ on the radiology dataset VQA-RAD~\cite{lau2018dataset}. Consistent with Kvasir-VQA-x1, \method{} achieves the best ROUGE-L and METEOR and the second-best BLEU while requiring only 64 unique images, confirming that density-aware multimodal calibration generalizes effectively to radiology VQA.}

\camera{Tab.~\ref{tab:efficiency} compares the hardware efficiency of BF16 and GPTQ across 200 Kvasir-VQA-x1 test queries on an NVIDIA A100 40GB GPU. W4A16 reduces the weight footprint by 73.4\%, substantially lowering memory requirements for clinical deployment. Meanwhile, end-to-end latency and generation throughput remain comparable to BF16: both models share the same unquantized 24-block Vision Transformer for high-resolution visual encoding, and W4A16 performs on-the-fly weight dequantization to 16-bit during inference to preserve clinical accuracy.}

\begin{table}[!t]
  \centering
  \caption{\camera{Comparison of calibration strategies on the VQA-RAD test set. Best and second-best results are shown in \textbf{bold} and \underline{underline}.}}
  \label{tab:vqarad}
  {\small
    \setlength{\tabcolsep}{4pt}
    \begin{tabular}{llccc}
      \toprule
      \textbf{\camera{Quantizer}} & \textbf{\camera{Calibration}} & \textbf{\camera{BLEU}} $\uparrow$ & \textbf{\camera{ROUGE-L}} $\uparrow$ & \textbf{\camera{METEOR}} $\uparrow$ \\
      \midrule
      \multicolumn{2}{l}{\camera{\textit{Half precision (BF16)}}} & \camera{0.4815} & \camera{0.6495} & \camera{0.3391} \\
      \midrule
      \multirow{4}{*}{\shortstack[l]{\camera{GPTQ} \\ \camera{(W4A16)}}}
      & \camera{COCO}             & \camera{0.4612}             & \camera{0.6220}             & \camera{0.3233}             \\
      & \camera{Random}           & \camera{\textbf{0.4706}}    & \camera{0.6249}             & \camera{0.3218}             \\
      & \camera{K-Means}          & \camera{0.4609}             & \camera{\underline{0.6295}} & \camera{\underline{0.3240}} \\
      & \camera{\method{} (Ours)} & \camera{\underline{0.4645}} & \camera{\textbf{0.6305}}    & \camera{\textbf{0.3277}}    \\
      \bottomrule
    \end{tabular}
  }
  \vspace{-6mm}
\end{table}

\begin{table}[!t]
  \centering
  \caption{\camera{Hardware efficiency of BF16 vs.\ GPTQ across 200 Kvasir-VQA-x1 test queries on an NVIDIA A100 40GB GPU. VRAM usage is reported in GB, latency in ms, and throughput in tokens/s.}}
  \label{tab:efficiency}
  \small{%
    \setlength{\tabcolsep}{4pt}%
    \begin{tabular}{lccc}
      \toprule
      \textbf{\camera{Model}} & \textbf{\camera{VRAM}} $\downarrow$ & \textbf{\camera{Latency}} $\downarrow$ & \textbf{\camera{Throughput}} $\uparrow$ \\
      \midrule
      \camera{Half precision} & \camera{8.46} & \camera{\textbf{4,393.37 $\pm$ 2,581.33}} & \camera{\textbf{3.21}} \\
      \camera{GPTQ (\method{})} & \camera{\textbf{2.25}} & \camera{4,495.79 $\pm$ 2,659.16} & \camera{3.14} \\
      \bottomrule
    \end{tabular}%
  }
  \vspace{-6mm}
\end{table}

\subsection{Discussion}

\revise{The results suggest that calibration sensitivity depends on how each quantizer uses its calibration inputs. MASQuant and QIG estimate activation-related statistics, smoothing factors, or token-level sensitivities from calibration examples, making distribution mismatch particularly relevant. GPTQ uses the examples mainly to estimate reconstruction statistics and is less affected, but it still benefits from a representative in-domain workload. This explains why the largest absolute improvement occurs with MASQuant while the same calibration strategy remains competitive across all three PTQ mechanisms.}

\revise{A second observation is that representativeness is more useful when it is defined through density rather than only geometric proximity. K-Means selects points near centroids, but medical image manifolds can contain heterogeneous visual modes and irregular densities. A centroid can be geometrically central without corresponding to a particularly representative real image. HDBSCAN instead identifies dense regions and selects real points with high membership probability. The ablation results are consistent with this distinction.}

\revise{Third, the QA expansion stage is essential to the VQA setting. Selecting representative images alone does not guarantee representative multimodal calibration. Because each image may support several questions, the same 23 images can generate 128 QA pairs with varied question formats, while round-robin sampling prevents frequent question classes from consuming the budget. \method{} therefore differs from a standard image core-set method: the selection target is a compact set of multimodal interactions anchored to representative visual content.}

\revise{From a deployment perspective, \method{} is attractive because it does not require modifying or retraining the VLM or the quantizer. The additional work occurs before PTQ, in feature extraction and density-based selection, after which the selected calibration set is consumed by the original quantization pipeline. This makes \method{} a lightweight data-centric component that can be integrated into existing compression workflows. \camera{In the Kvasir-VQA-x1 setting}, only 23 unique medical images are needed for 128 QA pairs, reducing the number of distinct images exposed to the calibration process. This may be operationally useful in privacy-sensitive environments, although \method{} itself does not provide a formal privacy guarantee.}

\revise{Finally, the evaluation highlights a limitation of lexical VQA metrics. BLEU, ROUGE-L, and METEOR quantify textual overlap and may not fully distinguish clinically equivalent paraphrases from clinically important semantic errors. The qualitative examples provide complementary evidence by illustrating errors in abnormality presence, lesion count, and lesion type. However, expert clinical evaluation would be required to establish whether the observed capability-retention improvements translate into improved clinical utility.}

\section{Conclusion}

\revise{We presented CoreCalib, a training-free and quantizer-agnostic framework for constructing compact calibration sets for resource-efficient medical vision-language models. By combining clinical question-class stratification, density-aware visual representative selection, and class-balanced QA expansion, CoreCalib consistently improves capability retention across MASQuant, GPTQ, and QIG under a fixed calibration budget. These results demonstrate that calibration-data construction is an important component of practical medical VLM compression.}

\revise{Future work will extend CoreCalib to joint image-text and model-activation spaces, additional medical VQA datasets, and more aggressive quantization settings. \camera{Beyond A100 profiling, we will assess latency, memory consumption, throughput, and energy efficiency on hospital workstations and resource-constrained edge devices.}}

\begin{credits}
\subsubsection{\ackname} 
This research is funded by Vietnam National University - Ho Chi Minh City (VNU-HCM) under Grant Number A2025-18-04.
\end{credits}

%
%
\bibliographystyle{splncs04}
\bibliography{references}
\end{document}